\documentclass[10pt,conference]{IEEEtran}
\IEEEoverridecommandlockouts
\usepackage{amsmath,amssymb,amsfonts}
\usepackage{algorithmic}
\usepackage{graphicx}
\usepackage{textcomp}
\usepackage{xcolor}
\def\BibTeX{{\rm B\kern-.05em{\sc i\kern-.025em b}\kern-.08em
    T\kern-.1667em\lower.7ex\hbox{E}\kern-.125emX}}

\usepackage[nolist]{acronym}
\usepackage{multirow}
\usepackage{tabularx}
\usepackage{booktabs}
\usepackage{subcaption}

\newcommand\blfootnote[1]{%
  \begingroup
  \renewcommand\thefootnote{}\footnote{#1}%
  \addtocounter{footnote}{-1}%
  \endgroup
}

\usepackage[backend=biber,style=numeric,citestyle=numeric,natbib=true,maxcitenames=6,maxbibnames=99]{biblatex}
\begin{document}

\begin{acronym}
\acrodef{PGD} {Projected Gradient Descent}
\acrodef{FGSM} {Fast Gradient Sign Method}
\acrodef{SPSA}{Simultaneous Perturbation Stochastic Approximation}
\acrodef{DSNGD}{Dynamically Sampled Nonlocal Gradient Descent}
\acrodef{ODI}{Output Diversified Initialization}
\acrodef{GD}{Gradient Descent}
\end{acronym}

\title{Dynamically Sampled Nonlocal Gradients\\ for Stronger Adversarial Attacks\\

}

\author{%
\IEEEauthorblockN{Leo~Schwinn\IEEEauthorrefmark{1}, An~Nguyen\IEEEauthorrefmark{1}, Ren\'e~Raab\IEEEauthorrefmark{1}, Dario Zanca\IEEEauthorrefmark{1}, Bjoern~M.~Eskofier\IEEEauthorrefmark{1}, Daniel Tenbrinck\IEEEauthorrefmark{2}, Martin Burger\IEEEauthorrefmark{2}} \\
\IEEEauthorblockA{\IEEEauthorrefmark{1}Machine Learning and Data Analytics Lab, Friedrich-Alexander Universit\"at Erlangen-N\"urnberg (FAU), Germany\\
  \texttt{\{leo.schwinn,an.nguyen,rene.raab,dario.zanca,bjoern.eskofier\}@fau.de}}\\

\IEEEauthorblockA{\IEEEauthorrefmark{2}Department Mathematics, Friedrich-Alexander Universit\"at Erlangen-N\"urnberg (FAU), Germany\\
  \texttt{\{daniel.tenbrinck,martin.burger\}@fau.de}}
  
}

\maketitle
\begin{abstract}

The vulnerability of deep neural networks to small and even imperceptible perturbations has become a central topic in deep learning research. Although several sophisticated defense mechanisms have been introduced, most were later shown to be ineffective. However, a reliable evaluation of model robustness is mandatory for deployment in safety-critical scenarios. 
To overcome this problem we propose a simple yet effective modification to the gradient calculation of state-of-the-art first-order adversarial attacks.
Normally, the gradient update of an attack is directly calculated for the given data point. This approach is sensitive to noise and small local optima of the loss function. Inspired by gradient sampling techniques from non-convex optimization, we propose Dynamically Sampled Nonlocal Gradient Descent (DSNGD). DSNGD calculates the gradient direction of the adversarial attack as the weighted average over past gradients of the optimization history. Moreover, distribution hyperparameters that define the sampling operation are automatically learned during the optimization scheme. We empirically show that by incorporating this nonlocal gradient information, we are able to give a more accurate estimation of the global descent direction on noisy and non-convex loss surfaces. In addition, we show that DSNGD-based attacks are on average $35\%$ faster while achieving $0.9\%$ to $27.1\%$ higher success rates compared to their gradient descent-based counterparts.

\end{abstract}

\section{Introduction}

\blfootnote{Accepted for publication at the IEEE International Joint Conference on Neural Networks (IJCNN) 2021 (DOI: 10.1109/IJCNN52387.2021.9534190).

\textcopyright 2021 IEEE. Personal use of this material is permitted. Permission from IEEE must be obtained for all other uses, in any current or future media, including reprinting/republishing this material for advertising or promotional purposes, creating new collective works, for resale or redistribution to servers or lists, or reuse of any copyrighted component of this work in other works.}

Deep learning has led to breakthroughs in various fields, such as computer vision \cite{Kaiming2016,krizhevsky2009} and language processing \cite{Oord2016}. Despite its success, it is still limited by its vulnerability to adversarial examples \cite{Szegedy2014}. In image processing, adversarial examples are small, typically imperceptible perturbations to the input that cause misclassifications. In domains like autonomous driving or healthcare this can potentially have fatal consequences. Since the weakness of neural networks to adversarial examples has been demonstrated, many methods were proposed to make neural networks more robust and reliable \cite{Goodfellow2015,Madry2018}. In a constant challenge between new adversarial attacks and defenses, most of the proposed defenses have been shown to be rather ineffective \cite{Carlini2019,Guo2018,Kurakin2018,Samangouei2018}. 

\begin{figure*}
    \centering
    \begin{subfigure}{0.19\textwidth}
      \includegraphics[width=\textwidth]{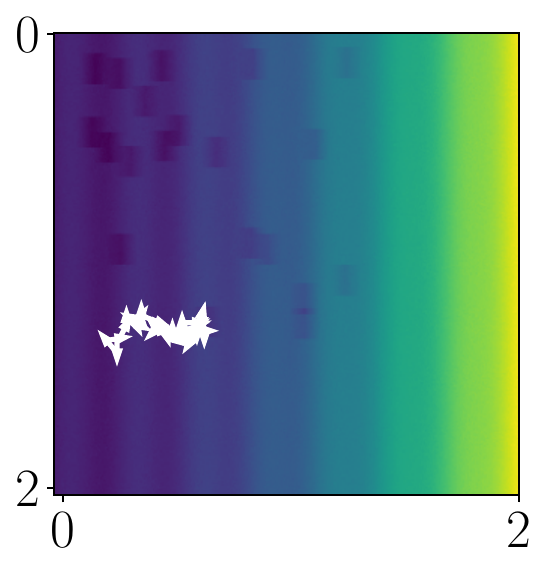}
      \caption{}
      \label{fig:gradient_descent}
    \end{subfigure} 
    \begin{subfigure}{0.19\textwidth}
      \includegraphics[width=\textwidth]{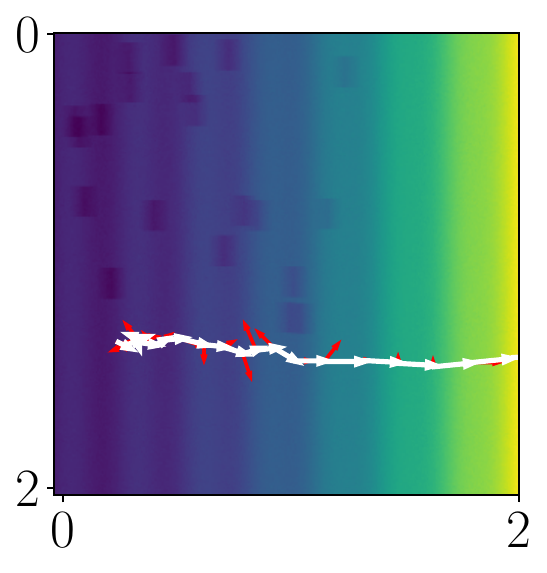}
      \caption{}
      \label{fig:gradient_descent_smoothed}
    \end{subfigure}
    \begin{subfigure}{0.22\textwidth}
      \includegraphics[width=\textwidth]{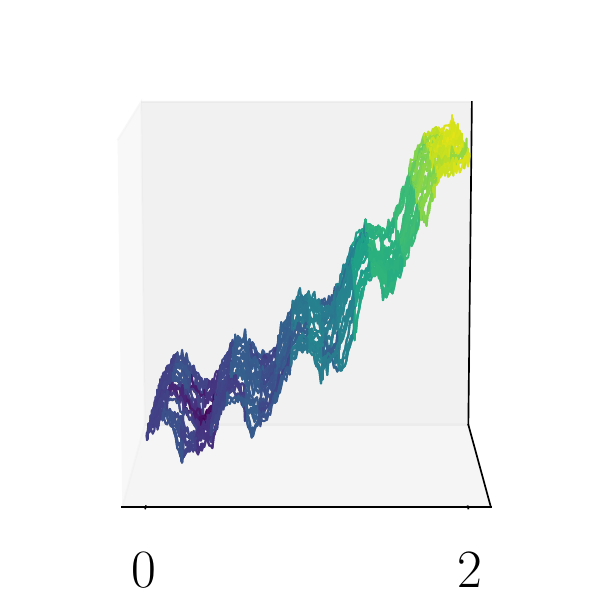}
      \caption{}
      \label{fig:surface}
    \end{subfigure} 
    \begin{subfigure}{0.22\textwidth}
      \includegraphics[width=\textwidth]{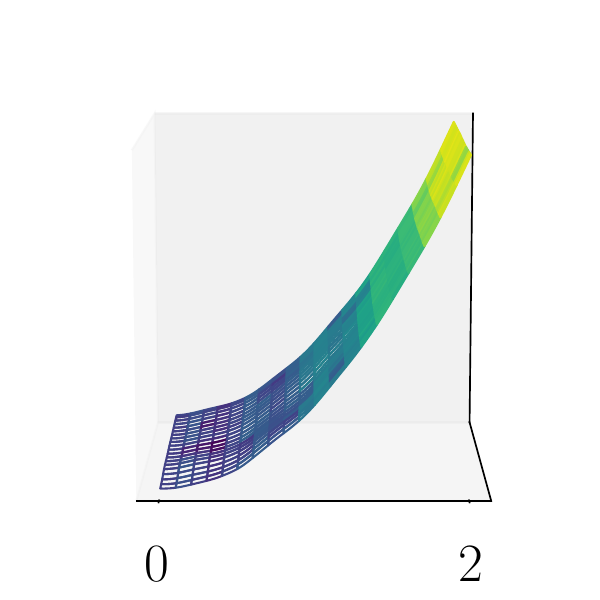}
      \caption{}
      \label{fig:surface_smoothed}
    \end{subfigure}
    \begin{subfigure}{0.065\textwidth}
      \includegraphics[width=\textwidth]{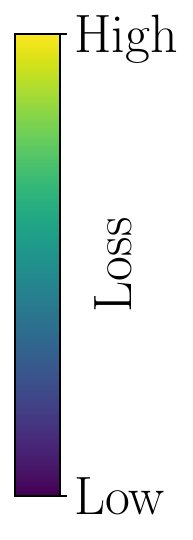}
      \label{fig:gradient_descent_cb}
      \caption*{}
    \end{subfigure}
  \caption{Comparison between the standard \acf{GD} method \textbf{(A)} and the proposed \acf{DSNGD} method \textbf{(B)}. \textbf{(C)} shows the side view of the noisy loss surface displayed in (A) and (B). In (A) and (B), the calculated ascent direction is displayed by a white arrow. For DSNGD the ascent direction is obtained by averaging the gradients of multiple samples (red arrows) in the history of the optimization. \textbf{(D)} illustrates the effect of DSNGD when the amount of sampling operations $N$ tends to infinity. Then, the mean of the sampled gradients corresponds to using GD on a loss surface that is convoluted with a convolution kernel dependent on the sampling distribution.}
  \label{gradient_descent_comparison}
\end{figure*}


\citet{Burke05} introduced the gradient sampling algorithm to reliably estimate the significant optima of non-smooth non-convex functions with unreliable gradient information. Gradient sampling can be interpreted as a generalized steepest descent method where the gradient is not only calculated at a given point but additionally for points in the direct vicinity. The final gradient direction is subsequently calculated in the convex hull of all sampled gradients and thus incorporates additional information about the local geometry of the loss surface. The central idea behind this approach is that point estimates of gradients can be misleading, especially for noisy and non-convex loss surfaces. However, a major limitation of this method is that explicitly computing the convex hull is not feasible for high-dimensional spaces. Recently, other sampling-based optimization algorithms have been proposed in the field of deep learning \cite{Athalye18Obfuscated,Athalye18EOT,Smilkov17,Wu2018}. \citet{Wu2018} introduced the variance-reduced attack (VRA) which approximates the gradient of the expected value as the average gradient over multiple noisy input samples. The goal of VRA is to improve the transferability of adversarial attacks between different models by reducing local oscillations of the loss surface. However, the calculation of the expected value over multiple input samples introduces a significant computational overhead dependent on the number of sampling operations. Furthermore, the estimated noise magnitude used for the sampling operation is an important hyperparameter that is difficult to tune in practice as it needs to be adjusted with respect to the respective optimization problem.



Inspired by gradient sampling algorithms from non-convex optimization and recent methods in the field of adversarial deep learning we aim to improve the effectiveness of adversarial attacks on deep neural networks. In our work we propose a general purpose \acf{GD} algorithm, named Dynamically Sampled Nonlocal Gradient Descent (DSNGD) and demonstrate its effectiveness on the optimization problem of adversarial attacks. We avoid the calculation of the convex hull and compute the final gradient direction as a weighted average over multiple gradients as in \cite{Athalye18EOT,Wu2018}. 
In contrast to prior work \cite{Athalye18EOT,Wu2018}, we keep a history of the already sampled data and the respective gradients to calculate the current gradient direction as their weighted average. In practice, this enables us to only sample once per gradient step while ultimately considering a similar amount of local information compared to sampling multiple times per gradient step. This effectively removes the computational overhead introduced by the sampling operation and makes the proposed algorithm orders of magnitude more efficient. Moreover, we directly learn the sampling distribution (and thus the size of the neighborhood) within the same optimisation scheme. Thus, we eliminate the need to manually tune an additional hyperparameter as necessary in prior \cite{Wu2018}. Fig.~\ref{gradient_descent_comparison} illustrates how the convergence of the standard GD method can be improved for noisy and non-convex loss surfaces by the proposed \acf{DSNGD} method. Standard GD calculates nearly random gradient directions while DSNGD approximates the global ascent direction more accurately as it is less susceptible to noise and local optima.

The contributions of this paper can be summarized as follows. First, we relate recently proposed sampling methods to the empirical mean, which converges towards a nonlocal gradient with respect to an underlying probability distribution with an increasing number of sampling operations. Next, we show that DSNGD-based adversarial attacks achieve higher success rates while being substantially more query-efficient compared to prior attacks on several benchmark datasets. Furthermore, we demonstrate that DSNGD approximates the direction of an adversarial example more accurately during the attack and we empirically show the effectiveness of DSNGD on non-convex loss surfaces. Finally, we demonstrate that DSNGD can optimally adapt the hyperparameters of the sampling procedure based on the characteristics of the loss landscape. 

\section{Preliminary}
In the following, we introduce the necessary mathematical notation to describe adversarial attacks and to review current contributions in this field.

\subsection{Notation}

Let ($x$, $y$) with $x\in\mathbb{R}^d$ and $y\in\{1, 2, \dots, C\}$ be pairs of samples (e.g., images) and labels in a classification task with $C$ different classes, where each sample is represented by a $d$-dimensional feature vector. We define a batch $M$ as a subset of ($x$, $y$) with $|M|$ tuples. In the following we assume that the samples are drawn from the $d$-dimensional unit cube, i.e., $x \in [0,1]^d$, as this is typically the case for image data. Let $\mathcal{L}$ be the loss function (e.g., categorical cross-entropy) of a neural network $F_{\theta}$, parameterized by the parameter vector $\theta \in \Theta$. 
Constructing an adversarial perturbation $\gamma \in \mathbb{R}^d$ with maximum effect on the loss value can be stated as the following optimization problem: 
\begin{align} \label{eq:adversarial_opt}
\underset{\gamma}\max ~\mathcal{L}(F_{\theta}(x + \gamma), y)
\end{align}
The perturbation $\gamma$ is usually constrained in two ways: 1) The value range of the adversarial example is still valid for the respective domain (e.g, between [$0$, $1$] or [$0$, $255$] for images), 2) The adversarial example $x_{adv} = x + \gamma$ is within a set of allowed perturbations $S$ that are unlikely to change the class label for human perception. In the following, we focus on \textit{untargeted gradient-based adversarial attacks} that are constrained by the $L_{\infty}$ norm such that $||\gamma||_{\infty} \leq \epsilon$, as done in prior work \cite{Lin2020}.

\subsection{Related Work}


The majority of adversarial attacks are gradient-based \cite{Goodfellow2015,Useato2018,Lin2020}. These attacks utilize the gradient information of the given model to construct optimal attacks for a given loss function and have shown a higher success rate while being more efficient than their non-gradient-based counterparts (e.g., decision-based \cite{Brendel18}, score-based \cite{Spall1992}). However, gradient-based attacks have shown to be futile if the model shows signs of obfuscated gradients \cite{Useato2018}, whereas non-gradient-based algorithms are not affected by gradient obfuscation \cite{Spall1992,Useato2018}. Recently, \citet{Brendel19} proposed a versatile gradient-based attack, named B\&B attack. The B\&B attack utilizes the gradient information of the model to estimate the decision boundary between an adversarial and a benign sample. This attack has shown to successfully attack models with obfuscated gradients. Other recent advances for adversarial attacks include better query efficiency by changing the optimization algorithm (e.g., to ADAM \cite{Useato2018}) or by finding more efficient starting points for the attacks \cite{tashiro2020}. In this work we consider the most popular and effective gradient-based algorithms including Projected Gradient Descent (PGD) \cite{Madry2018} and its variants \cite{Lin2020}, the B\&B attack \cite{Brendel19}, and a proven score-based attack SPSA \cite{Spall1992} for a diverse set of attacks.

\subsection{Sampled Nonlocal Gradients}

Recently developed methods in the field of deep learning aim to improve gradient information by averaging multiple sampled gradients in the vicinity of a data point \cite{Athalye18Obfuscated,Athalye18EOT,Smilkov17,Wu2018}. These methods can be related to Sampled nonlocal gradients (SNG). SNG are specifically designed for noisy and non-convex loss surfaces as they are calculated as the weighted average over multiple sample points in the vicinity of the current data point. The gradient calculation of a sampled nonlocal gradient is given by:
\begin{equation} \label{eq:gradient_averaging}
\begin{split}
\nabla_{x}^{\scriptscriptstyle{SNG}} &\mathcal{L}(F_\theta (x),y) := \nabla_x \frac{1}{N} \sum_{i=1}^{N} \mathcal{L}(F_\theta (x + \xi_i^\sigma), y),
\end{split}
\end{equation}
where $N\in\mathbb{N}$ is the number of sampling operations. The random variables $\xi_i^\sigma$ are considered to be drawn i.i.d. from a distribution $P^\sigma$ parametrized by the standard deviation $\sigma  \in \mathbb{R}$ which effectively determines the size of the neighborhood. By the law of large numbers, the sampled nonlocal gradient converges (with a rate of order $N^{-1/2}$ in variance) to the respective expected value, which is given by the nonlocal gradient
\begin{align} 
\nabla_x \mathbb{E}_{\xi \sim P^\sigma} \left[ \mathcal{L}(F_\theta (x +\xi), y)  \right].
\end{align}
The expectation is effectively a local averaging of the likelihood around $x$ (cf. Fig. \ref{fig:gradient_descent_smoothed}. Note that by linearity, the SNG is equivalent to an averaging of the gradients, i.e., the standard form of a nonlocal gradient, cf. \cite{du2019nonlocal}. Although sampled nonlocal gradients are an effective way to estimate the descent direction for non-smooth and non-convex loss surfaces, the sampling operation introduces a substantial computational overhead.



\section{Dynamically Sampled Nonlocal Gradient Descent} \label{SampledNonlocal}

We propose an efficient approximation for sampled nonlocal gradients, which we embed into a novel algorithm named \acf{DSNGD}. DSNGD calculates the gradient direction as the weighted average over the optimization history of the current and past gradients. Instead of using multiple sampling operations in every attack iteration, we only add one sample to the optimization history in every iteration. Thus, a large number of samples can be used to calculate gradient directions without incurring noticeable computational overhead. The gradient calculation of DSNGD is given by:
\begin{equation} \label{eq:DSNG}
\begin{split}
\nabla_{x}^{\scriptscriptstyle{DSNGD}} &\mathcal{L}(F_\theta (x_t),y) := \sum_{i = 1}^{t}  w_{i} \cdot \nabla_{x} \mathcal{L}(F_\theta (\hat{x}_i), y) \\
 \text{where } & \hat{x}_i = \operatorname{clip_{[0,1]}}\{x_i + \xi_i^\sigma\},
\end{split}
\end{equation}
where $\hat{x}_i$ are the noisy samples in the optimization history and $w_i$ is the gradient weight associated to $\hat{x}_i$. Furthermore, $t$ describes the current attack iteration. Note that we store past gradients in memory rather than recalculating them in every attack iteration, which requires only a neglectable amount of memory. $\operatorname{clip_{[a,b]}}$ is the component-wise clipping operator with value range $[a, b]$. The clipping operator is needed to ensure that the data stays in the normalized range, e.g., in the case of images. It can be discarded for other applications with unbounded data. Similar to SNG, the random variables $\xi_i^\sigma$ are considered to be drawn i.i.d. from a distribution $P^\sigma$ parametrized by the standard deviation $\sigma  \in \mathbb{R}^{|M|}$. Here $\sigma_m$ is the $m$-th entry of $\sigma$ where $\sigma_m > 0, \forall m$. We optimize each $\sigma_m$ individually for every sample in a batch $M$ during the respective attack. This is achieved by calculating the gradient with respect to $\sigma$ within the attack optimization via backpropagation. The gradient update with respect to $\sigma$ can be formalized as
\begin{equation}
    \sigma_{t+1} = \sigma_{t} - \lambda_{\sigma} \cdot \nabla_{\sigma} \mathcal{L}(F_\theta (\operatorname{clip_{[0,1]}}\{x_t + \xi_t^\sigma\}), y),
\end{equation}
where $\lambda_{\sigma}$ is the learning rate of the distribution hyperparameter $\sigma$.

Using the definition of the DSNGD gradient introduced in \eqref{eq:DSNG}, we are able to demonstrate the simplicity of incorporating our method in other gradient-based methods with the example of PGD attacks. An DSNGD-based PGD attack is formally given by
\begin{align} \label{eq:sngdifgsm}
    x_{t + 1} = \Pi_{S} \, (x_{t} + \alpha \cdot \operatorname{sign}(\nabla_{x_t}^{\scriptscriptstyle{DSNGD}}  \mathcal{L}(F_\theta(x_{t}), y)),
\end{align}
where $\Pi_{S}()$ is a projection operator that keeps the perturbation $\gamma$ within the set of valid perturbations $S$ and $\alpha$ is the step size of the attack. We start the iterative update with $x_0 = x$.

\section{Experiments}

We conduct several experiments to evaluate DSNGD. We first analyze if DSNGD can improve the success rate of adversarial attacks compared to GD. Secondly, we inspect the performance difference between several attacks, as efficient attacks play an important role in facilitating the evaluation of model robustness in real-world applications. Furthermore, we explore the possibility of combining DSNGD with other methods, like Output Diversified Initialization (ODI) \cite{tashiro2020}, which is designed to improve the starting point of an attack. Lastly, we analyze the ability of DSNGD to better approximate the global descent direction and show that DSNGD is effective for non-convex loss surfaces.

\subsection{Setup} 

In the following we give an overview of general hyperparameters used for the experiments, including threat model, training, evaluation, and datasets. We describe dataset-specific hyperparameters such as the model architecture in the corresponding sections.

\subsubsection{Threat Model} \label{Thread_model}In this work we focus our evaluation on the $L_{\infty}$ norm and untargeted attacks. We combine our proposed method, DSNGD, with state-of-the-art attacks, including PGD \cite{Madry2018} and PGD with Nesterov Momentum (N-PGD) \cite{Lin2020}, which we call DSN-PGD and DSN-N-PGD respectively. We additionally combine all PGD-based attacks with ODI \cite{tashiro2020}. ODI-based attacks achieve one of the highest success rates on the Madry MNIST leaderboard \cite{Madry2018}. Moreover, we compare our approach to the B\&B attack \cite{Brendel19}, one of the most recent and effective gradient-based attacks. We additionally evaluated if models are obfuscating their gradients with the zeroth order \ac{SPSA} attack \cite{Spall1992,Useato2018}.


We limited the amount of model evaluations to $2000$ for each gradient-based attack as we generally did not observe any success rate increases for more evaluations. We distributed the model evaluations over multiple restarts, such that $R\cdot I = 2000$, where $R$ denotes the total number of restarts and $I$ the amount of attack iterations. We tried multiple combinations of model evaluations and restarts, as we observed that for a fixed budget of total evaluations this has considerable impact on the performance of the attacks. For the standard PGD attack we obtained the highest success rates between $20-400$ iterations and $5-100$ random restarts. For DSN-PGD attacks we used $400$ iterations and $5$ random restarts. For both DSN-PGD and PGD we explored step sizes between $\alpha = \frac{1}{40} \epsilon$ and $\alpha = \frac{1}{2} \epsilon$. For the momentum-based attacks, we considered momentum values between $0.5$ and $1.0$. We used the AdverTorch \cite{Ding2019} implementation of the B\&B attack and performed it with $2000$ iterations. We considered learning rates between $0.0001$ and $0.01$ for the B\&B attack. The SPSA attack was performed with $100$ steps and a sample size of $8192$, as shown to be effective in \cite{Useato2018}. We used $2$ ODI steps to warmstart any ODI enhanced adversarial attack as proposed in the original paper \cite{tashiro2020}. We performed all attacks on the same subset of $1000$ ($10\%$) randomly selected test images as in \cite{Brendel19,Useato2018}.

\subsection{Data and Architectures}

Three different image classification datasets were used to evaluate the efficiency of the adversarial attacks (MNIST \cite{LeCun98}, Fashion-MNIST \cite{Xiao2017} and CIFAR10 \cite{krizhevsky2009}). We split each dataset into the predefined train and test sets and additionally removed $10\%$ of the training data for validation.

\subsubsection{Training} We evaluated our attack on two of the strongest empirical defenses to date, adversarial training \cite{Athalye18Obfuscated,Madry2018,Useato2018} and TRADES \cite{Zhang19}. We additionally explore a defense based on restricting the hidden space of neural networks (HSR) \cite{Mustafa19}, that has shown to be robust against PGD but has proven to be ineffective against stronger attacks \cite{Croce2020}. For adversarial training we used the fast-FGSM-based adversarial training algorithm \cite{Wong2020}. In preliminary experiments on MNIST we observed that the loss surfaces of these models are not as convex as described in prior work \cite{Kurakin2018} (see Fig.~\ref{fig:loss_landscape_fgsm_cifar}) compared to models trained with PGD-based adversarial training. For comparison we additionally trained each model with the typically used PGD-based adversarial training \cite{Madry2018}. To evaluate the methods proposed in \cite{Zhang19} and \cite{Mustafa19} we used the pre-trained models provided by the authors. For fast-FGSM-based training we used the same hyperparameters as proposed in \cite{Wong2020}. For PGD-based training we used $7$ steps and a step size of $1/4$ $\epsilon$ \cite{Madry2018}. All self-trained networks were trained and evaluated $5$ times using stochastic gradient descent with the Adam optimizer ($\beta_{1} = 0.9$, $\beta_{2} = 0.999$) \cite{Kingma14}. We used a cyclical learning rate schedule \cite{Smith2017}, which has been successfully used for adversarial training in prior work \cite{Wong2020}. Thereby, the learning rate $\lambda$ was linearly increased up to its maximum $\Lambda$ over the first $2/5$ epochs and then decreased to zero over the remaining epochs. The maximum learning rate $\Lambda$ was estimated by increasing the learning rate of each individual network for a few epochs until the training loss diverged \cite{Wong2020}. All models were optimized for $100$ epochs, which was sufficient for convergence. The checkpoint with the lowest adversarial validation loss was chosen for testing.

\subsubsection{MNIST} consists of greyscale images of handwritten digits each of size $28\times28\times1$ ($60,000$ training and $10,000$ test). We used the same MNIST model as in \cite{Wong2020}. However, we doubled the number of filters for the convolutional layers, as we noticed that the performance of the model sometimes diverged to random guessing during training. As in prior work, we used a maximum perturbation budget of $\epsilon = 0.3$.

%




\subsubsection{Fashion-MNIST} consists of greyscale images of $10$ different types of clothing, each of size $28\times28\times1$ ($60,000$ training and $10,000$ test). The Fashion-MNIST classification task is slightly more complicated than MNIST, as it contains more intricate patterns. We used the same model architecture as for MNIST. The optimal learning rate we found for Fashion-MNIST was approximately $0.007$. To the best of our knowledge there is no standard perturbation budget $\epsilon$ commonly used for Fashion-MNIST. Since this dataset contains more complicated patterns than MNIST we used a lower maximum perturbation budget of $\epsilon = 0.15$.



\subsubsection{CIFAR10} consists of color images, each of size $32\times32\times3$, with $10$ different labels ($50,000$ training and $10,000$ test). CIFAR10 is the most challenging classification task out of the three. For CIFAR10 we used the same PreActivationResNet18 \cite{Kaiming2016} architecture as in \cite{Wong2020}. All images from the CIFAR10 dataset were standardized and random cropping and horizontal flipping were used for data augmentation during training as in \cite{Kaiming2016,Madry2018,Wong2020}. We found the optimal learning rate to be around $0.21$. In line with previous work, we set the maximum perturbation budget to $\epsilon = 8/255$.

\subsection{Experiments on DSNGD Parameters}

\begin{table*}
    \centering
    \begin{tabular}{lcccccccc}
        \toprule
        & Clean & FGSM & PGD & N-PGD &B\&B & DSN-PGD & DSN-N-PGD & SPSA \\
        \midrule
        \textbf{MNIST} \\
        fast-FGSM & 99.2 \textsubscript{\textpm 0} & 96.4 \textsubscript{\textpm 2} & 88.8 \textsubscript{\textpm 3} & 88.6 \textsubscript{\textpm 4} & 86.8 \textsubscript{\textpm 3} & \textbf{85.0}\textsubscript{\textpm 2} & 85.2 \textsubscript{\textpm 1} & 92.8\textsubscript{\textpm 3}\\
        PGD & 99.0 \textsubscript{\textpm 0} & 97.0 \textsubscript{\textpm 2} & 92.4 \textsubscript{\textpm 2} & 93.6 \textsubscript{\textpm 2} &  90.8 \textsubscript{\textpm 3} & \textbf{90.1}\textsubscript{\textpm 1} & \textbf{90.1} \textsubscript{\textpm 2} & 95.0 \textsubscript{\textpm 3} \\
        TRADES \cite{Zhang19} & 99.5 & 96.2 & 91.2 & 91.6 & 90.6 & \textbf{89.9}  & 90.0 & 92.2 \\
        \midrule
        \textbf{F-MNIST} \\
        fast-FGSM & 85.4 \textsubscript{\textpm 1} & 74.8 \textsubscript{\textpm 4} & 60.4 \textsubscript{\textpm 8} & 61.6  \textsubscript{\textpm 8}  & 60.5  \textsubscript{\textpm 8} & \textbf{58.4}\textsubscript{\textpm 8} & 59.6 \textsubscript{\textpm 8} & 66.6\textsubscript{\textpm 8}\\
        PGD &  85.7 \textsubscript{\textpm 0} &  83.2 \textsubscript{\textpm 5} & 70.0 \textsubscript{\textpm 7} & 69.2  \textsubscript{\textpm 7} & 70.2 \textsubscript{\textpm 6} & 67.6 \textsubscript{\textpm 5} & \textbf{67.4} \textsubscript{\textpm 6} & 74.2 \textsubscript{\textpm 6} \\
         \midrule
        \textbf{CIFAR10} \\
        fast-FGSM & 83.6 \textsubscript{\textpm 0} & 54.0 \textsubscript{\textpm 7} & 43.4 \textsubscript{\textpm 6} & 44.0 \textsubscript{\textpm 4} & \textbf{37.0} \textsubscript{\textpm 15} & \textbf{37.0} \textsubscript{\textpm 13} & 39.0\textsubscript{\textpm 12}  & 49.1\textsubscript{\textpm 6}\\
        PGD & 79.7 \textsubscript{\textpm 0} & 55.0 \textsubscript{\textpm 3} & 48.4 \textsubscript{\textpm 2} & 49.6 \textsubscript{\textpm 2}  & 48.6 \textsubscript{\textpm 3} &\textbf{47.0} \textsubscript{\textpm 2} & 47.2 \textsubscript{\textpm 1} & 50.5 \textsubscript{\textpm 4} \\
        TRADES \cite{Zhang19}  & 84.9 & 63.2 & 59.3 & 59.2 & 58.3 & \textbf{58.0} & 58.4 & 60.2 \\
        HSR \cite{Mustafa19} & 89.3 & 60.2 & 27.1 & 32.4 & \textbf{0.0} & \textbf{0.0} & 12.3 & 42.1 \\
        
    \end{tabular}
     \caption{Mean accuracy and standard deviation ($\%$) for various adversarial attacks. The attack with the highest success rate is displayed in bold for each row (lowest accuracy). fast-FGSM- and PGD-trained models were trained and evaluated five times.}
     \label{tab:first_results}
\end{table*}

\subsubsection{Noise Distribution} To combine \ac{DSNGD} with an adversarial attack, we need to define the distribution $P^\sigma$ from which we sample data points. This distribution should sample data points on the manifold of possible solutions of the optimization problem to provide informative gradients. For example, this can be a uniform distribution for the $L_\infty$-norm (sampling in the hypercube) or a Gaussian distribution for the $L_2$-norm (sampling in the hypersphere). For other optimization problems, e.g., finding an adversarial rotation that will result in a misclassification, one could sample random rotations within a learnable angle. Since we performed all attacks in the $L_\infty$ norm, we used a uniform distribution for all experiments.
The optimal standard deviation $\sigma$ of the distribution differs depending on the optimization problem. In our experiments, large values of $\sigma$ degrade the performance while very small values of $\sigma$ have no considerable impact. Instead of manually tuning this hyperparameter we learn it during the attack optimization. We constrained the search space for the optimal standard deviation $\sigma$ to $0< \sigma <\epsilon$ since the gradient information outside of the attack radius should be non-relevant for the optimization of the attack. We started with $\sigma = \epsilon$ and set $\lambda_{\sigma} = 1$ for all experiments.

\subsubsection{Sampling} 
\begin{figure}[t]
\centering
 \begin{subfigure}{0.24\textwidth}
    \includegraphics[width=\textwidth]{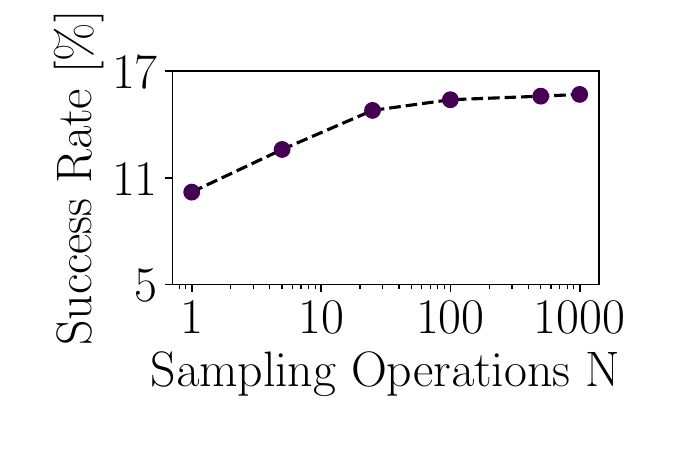}
    \caption{}
 \end{subfigure}
  \begin{subfigure}{0.24\textwidth}
    \includegraphics[width=\textwidth]{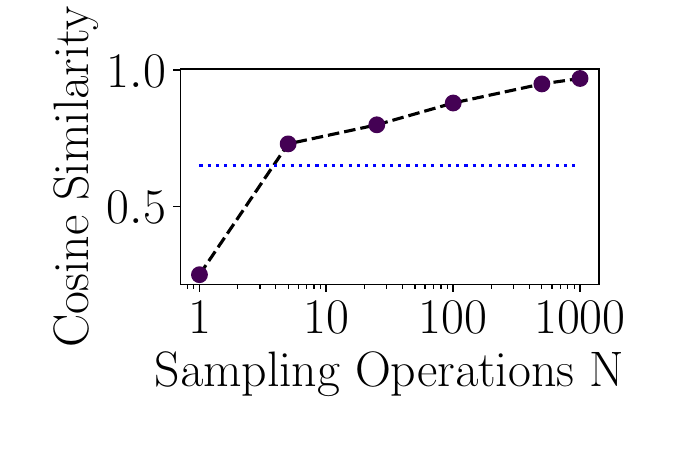}
    \caption{}
 \end{subfigure}
\caption{Average attack success rate \textbf{(A)} and cosine similarity \textbf{(B)} between subsequent gradient directions of a SNG-based PGD attack for varying amounts of sampling operations $N$ for the MNIST dataset (x-axis is on a logarithmic scale). The dotted blue line indicates the average cosine-similarity for a PGD-based attack.}
\label{fig:cosinesim_subsequent}
\end{figure}

In a preliminary experiment we evaluated if the performance of adversarial attacks increases by using sampled nonlocal gradients (SNGs). Therefore, we replaced the gradient of a PGD attack with the SNG in \eqref{eq:gradient_averaging} and evaluated the performance with an increasing amount of sampling operations $N$ on the MNIST validation set. Each sampling operation increases the computational overhead of SNG-based attacks but should in turn improve the calculated descent direction. We observed that the success rate monotonically increased for large amounts of sampling operations until it saturates at around $N=500$ (see Fig.~\ref{fig:cosinesim_subsequent}). In practice, the amount of sampling operations is limited by the computation budget. This makes this approach infeasible in practice, as the computational effort increases substantially with the amount of sampling operations. 

\subsubsection{Weighting}

We expect gradients of inputs that have a large distance to the current data point to be less relevant to the optimization. In line with this intuition, we observed that naively averaging the gradients results in weak attacks and propose to weight them according to their input distance. Since we focus on $L_{\infty}$ norm attacks, we consider the $L_{\infty}$ norm as the distance measure between data points. The weights are set based on the following rule:

\begin{equation*}
  w_{i}=\begin{cases}
    \exp({-\beta ||x_t - \hat{x}_i||_{\infty}}), & \text{if $||x_t - \hat{x}_i||_{\infty} > \sigma$}.\\
    1, & \text{otherwise}.
  \end{cases}
\end{equation*}

where $x_t$ is the current data point and $\hat{x}_i$ is the noisy input of the $i$-th attack iteration. This results in a lower importance of the gradients $\nabla \hat{x}_i$ where the associated input $\hat{x}_i$ has an $L_{\infty}$ norm distance to the current adversarial example greater than $\sigma$. At the same time, input samples which are still within the uniform sampling distribution defined by $\sigma$ are assigned a large weight. We did not observe considerable differences for $\beta$ values between $5$ and $20$ and set $\beta = 10$ for all experiments.

\section{Results and Discussion} \label{Sec:Results}

\subsection{Success Rate} Table \ref{tab:first_results} demonstrates that the proposed DSN-PGD attack surpasses the mean success rate of prior attacks in our experiments in $7$ out of $9$ cases.
No consistent performance increase can be observed when combining any of the PGD-based methods with Nesterov momentum. Nesterov PGD (N-PGD) outperforms PGD in $3$ out of $9$ cases and SN-Nesterov-PGD (DSN-N-PGD) outperforms DSN-PGD in $1$ out of $9$ cases. We additionally analyze the performance for individual runs as the standard deviation is high in some cases (e.g., F-MNIST). For all experiments where DSN-PGD shows the highest mean success rate it also shows the highest success rate for all individual models. DSN-PGD showed equal performance compared to the B\&B attack for two models. This includs one of the models trained with fast-FGSM (CIFAR10) and the pre-trained model of \cite{Mustafa19}. Both models exhibit low accuracy against the B\&B and DSN-PGD attack but showe robustness against the other attacks including SPSA. We observed that these model showed signs of obfuscated gradients (see loss surface in Fig.~\ref{fig:loss_landscape_fgsm_cifar}). These results indicate that the DSN-PGD attack is more reliable for models with obfuscated gradient information than standard PGD. 

\subsection{Runtime Comparison}

Table \ref{tab:runtime} shows the runtime average and standard deviation for each attack to achieve standard PGD performance. The values are calculated over all experiments shown in Table \ref{tab:first_results}. Due to the lack of gradient information, SPSA requires a high amount of model evaluations. The SPSA attack does not achieve PGD performance in our experiment with $400$ times the computational budget. The Foolbox \cite{rauber2017} implementation provided by the authors of the B\&B attack \cite{Brendel19} is also considerably slower than PGD for the same amount of model evaluations in our experiments. DSN-PGD achieves the same success rate compared to PGD considerably faster while at the same time achieving higher success rates with the same amount of model evaluations. The runtimes of the attacks were compared on a Nvidia GeForce GTX1080.

\begin{table}
    \centering
    \begin{tabular}{lcccc}
        \toprule
        & PGD & DSN-PGD & B\&B & SPSA \\
        \midrule
         Time & 100 \textsubscript{\textpm 4.9} & \textbf{74} \textsubscript{\textpm 1}  & 389  \textsubscript{\textpm 23}  & N/A \\
    \end{tabular}
    \caption{Mean and standard deviation of the relative runtime [\%] of several attacks to achieve equal performance on all datasets compared to standard PGD (100\%).}
    \label{tab:runtime}
\end{table}


\subsection{Additional Experiments}

\begin{figure*}[t]
 \centering
    \begin{subfigure}{0.48\textwidth}
      \includegraphics[width=0.54\textwidth]{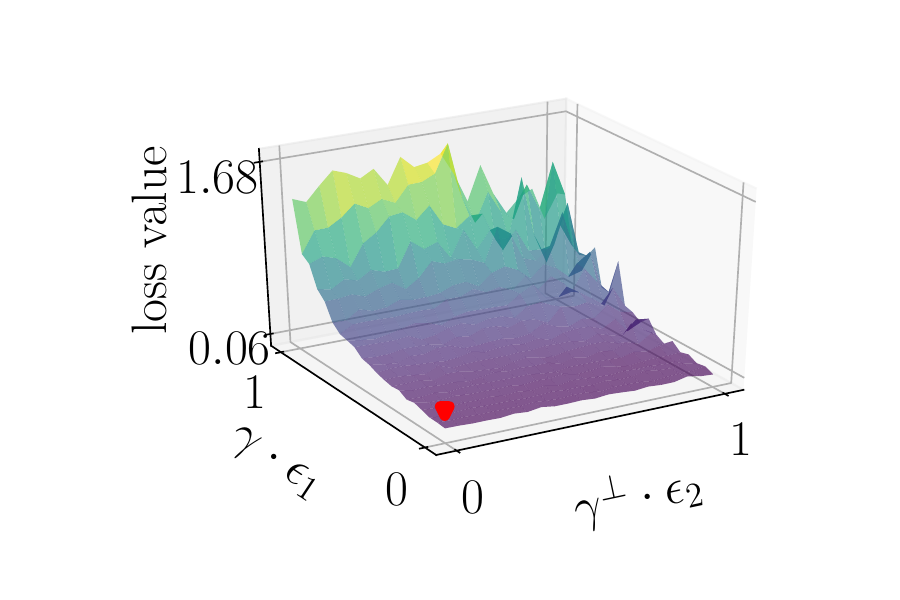}
      \includegraphics[width=0.45\textwidth]{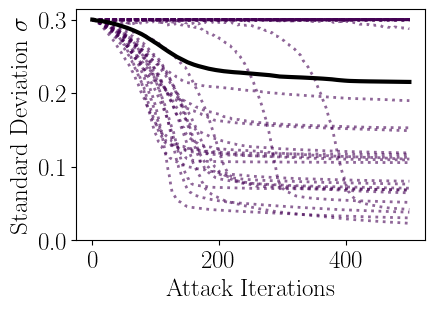}
      \caption{FGSM CIFAR10 loss landscape and noise behavior}
      \label{fig:loss_landscape_fgsm_cifar}
    \end{subfigure} 
    \begin{subfigure}{0.48\textwidth}
      \includegraphics[width=0.54\textwidth]{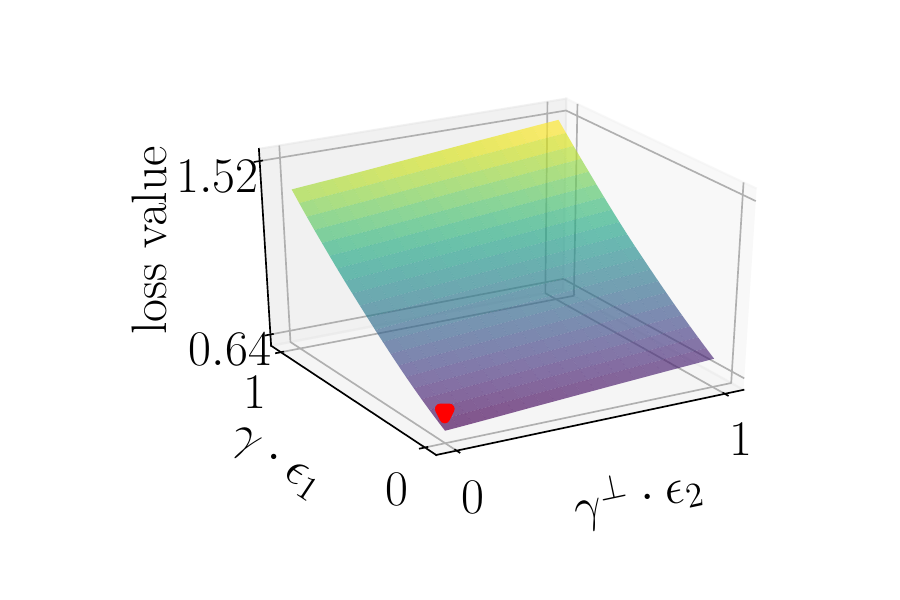}
      \includegraphics[width=0.45\textwidth]{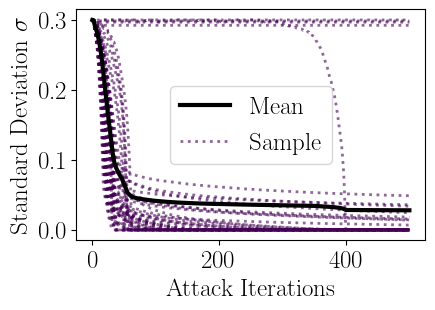}
      \caption{PGD CIFAR10 loss landscape and noise behavior}
      \label{fig:loss_landscape_cifar}
    \end{subfigure} 
 \caption{Representative loss surfaces around a clean sample $X_i$ (red dot) for a fast-FGSM-trained model \textbf{(A)} and a PGD-trained model \textbf{(B)}. We calculate the loss value for sample $x_{i} + \epsilon_{1} \cdot \gamma + \epsilon_{2} \cdot \gamma^\perp$ where $\gamma$ is the direction of a successful adversarial attack and $\gamma^\perp$ a random orthogonal direction. In addition, the behavior of $\sigma$ for individual samples throughout the attack optimization of a single batch is shown.}
 \label{fig:loss_landscape}
\end{figure*}

The following section summarizes additional experiments: 1) combination of DSNGD with ODI, 2) the ability to approximate the global descent direction between GD- and DSNGD-based attacks, and 3) the effectiveness of DSNGD on increasingly convex loss surfaces. 


\subsubsection{Combination with ODI} We evaluate if combining the different PGD-based attacks with \acf{ODI} increases their success rate. Combining ODI with PGD yields the same benefit as combining it with DSN-PGD in our experiments. On the CIFAR10 dataset, initialization with ODI yields the biggest improvement for PGD and DSN-PGD with $1.8\%$ and $2.2\%$ increase in success rate respectively. For MNIST and Fashion-MNIST, the performance was not changed considerably. This is in line with the original paper, where the success rate increased only marginally on the MNIST dataset and more substantially on CIFAR10 \cite{tashiro2020}. The results are summarized in Table \ref{tab:odi_results}. The results demonstrate that it can be beneficial to combine DSN-PGD with other methods. Note that ODI is partly designed to increase the transferability of adversarial attacks to other models which was not tested in this experiment. 

\begin{table}[t]
     \centering
     \begin{tabular}{lrrr}
         \toprule
         ODI & PGD & N-PGD & DSN-PGD \\
         \midrule
       \textbf{MNIST} \\
         FGSM & 88.6\textsubscript{\textpm 3} [-0.2] & 88.4\textsubscript{\textpm 3} [-0.2] & \textbf{84.9}\textsubscript{\textpm 2} [-0.1] \\
         PGD & 92.4\textsubscript{\textpm 2} [-0.0] & 93.3\textsubscript{\textpm 2} [-0.3] & \textbf{90.0}\textsubscript{\textpm 2} [-0.1]  \\
         TRADES \cite{Zhang19} & 91.2 [-0.0] & 91.4 [-0.2] & \textbf{89.8} [-0.1] \\
         \midrule
         \textbf{F-MNIST} \\
         FGSM & 60.4\textsubscript{\textpm 8} [-0.0]  & 61.4\textsubscript{\textpm 7} [-0.2] & \textbf{58.3}\textsubscript{\textpm 8} [-0.1] \\
         PGD & 70.0\textsubscript{\textpm 7} [-0.0]  & 69.1\textsubscript{\textpm 7} [-0.1] & \textbf{67.6}\textsubscript{\textpm 5} [-0.0]  \\
         \midrule
         \textbf{CIFAR10} \\
        FGSM & 42.3\textsubscript{\textpm 6} [-1.1]  & 42.7\textsubscript{\textpm 3} [-1.3] & \textbf{35.7}\textsubscript{\textpm 14} [-1.3]  \\
         PGD & 47.2\textsubscript{\textpm 2} [-1.2]  & 48.2\textsubscript{\textpm 2} [-1.4]  & \textbf{44.8}\textsubscript{\textpm 1} [-2.2]  \\
         TRADES \cite{Zhang19} & 57.5 [-1.8] & 57.7 [-1.5] & \textbf{56.3} [-1.7] \\
         HSR \cite{Mustafa19} & 25.1 [-2.0] & 30.5 [-1.9] & \textbf{0.0} [-0.0]
     \end{tabular}
     \caption{Mean accuracy and standard deviation ($\%$)  for various adversarial attacks with ODI. The performance difference to attacks without ODI is given by a subscript (negative subscript values indicate an attack success rate increase). The attack with the highest success rate is displayed in bold for each row. FGSM- and PGD-trained models were trained and evaluated five times.}
     \label{tab:odi_results}
\end{table}

\subsubsection{Approximation of the Adversarial Direction} 
\begin{figure}[t]
\centering
 \begin{subfigure}{0.24\textwidth}
    \includegraphics[width=\textwidth]{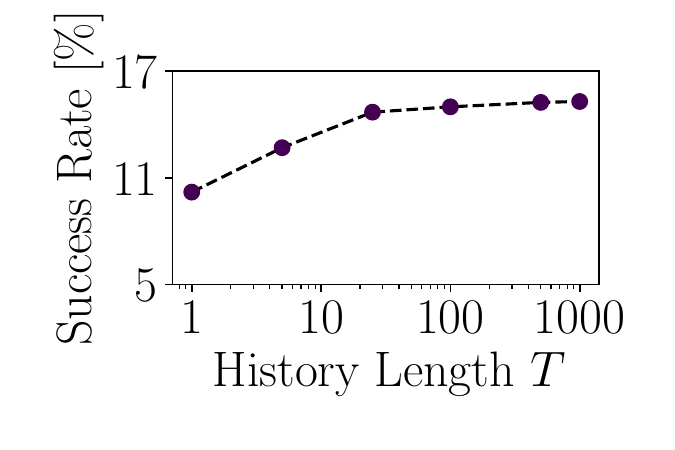}
    \caption{}
 \end{subfigure}
  \begin{subfigure}{0.24\textwidth}
    \includegraphics[width=\textwidth]{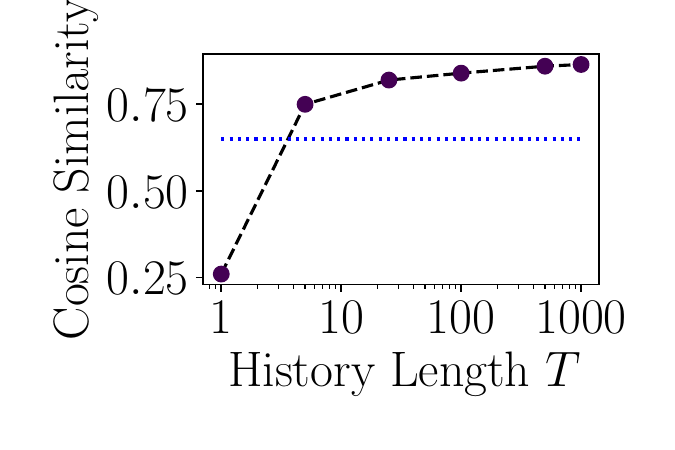}
    \caption{}
 \end{subfigure}
\caption{Average attack success rate \textbf{(A)} and cosine similarity \textbf{(B)} between subsequent gradient directions of a DSN-PGD attack for varying length of the optimization history $T$ for the MNIST dataset (x-axis is on a logarithmic scale). The dotted blue line indicates the average cosine-similarity for a PGD-based attack.}
\label{fig:cosinesim_subsequent_history}
\end{figure}

To get a better understanding of the effectiveness of DSNGD, we inspect if \ac{DSNGD}-based PGD attacks approximate the final direction of a successful adversarial attack more accurately. This is achieved by computing the average cosine similarity between subsequent iterations of the attack. Fig.~\ref{fig:cosinesim_subsequent_history} shows that increasing the total maximum length of the optimization history $T$ simultaneously increases the cosine similarity between subsequent attack iterations. Limiting the history length reduces the success rate of the attack and the average cosine similarity between attack iterations. The observed effect with a history length of $T = 1000$ is approximately equivalent to using $N = 100$ sampling operations in every attack iteration as displayed in Fig.~\ref{fig:cosinesim_subsequent}. The averaging of multiple gradients effectively approximates a smoothing of the loss function of the neural network $F_{\theta}$. The smoothing removes specific artifacts and minor local minima present in the loss landscape. Furthermore, it is automatically adapted to the optimization problem through the learnable hyperparameter $\sigma$ of the sampling distribution. Thus, the final adversarial direction is estimated more accurately by the \ac{DSNGD}-based attack compared to traditional PGD attacks.

\subsubsection{Loss Surface} After observing that DSNGD can improve the approximation of the global descent direction, we examined if DSNGD has a bigger impact on models with non-convex loss surfaces, where the global descent direction is hard to approximate for standard GD. Therefore, we construct an approximate visualization of the loss landscape by calculating the loss value along the direction of a successful adversarial perturbation ($g$) and a random orthogonal direction ($g^\perp$) originating from a clean sample as exemplified in Fig.~\ref{fig:loss_landscape}.
In contrast to prior work \cite{Kurakin2018} we find that the loss surface of the adversarially trained models (fast-FGSM and PGD) is often not increasing most rapidly towards the adversarial direction, which shows the non-convexity of the optimization problem. Furthermore, we noticed that in cases where the loss surface is more convex, the performance difference between PGD and DSN-PGD decreases. This behaviour is exemplified in the sub-figures \textbf{(2A)} and \textbf{(2B)}. The shown loss surfaces are increasingly convex, simultaneously the performance difference between DSN-PGD and PGD for these models decreases. Where DSN-PGD achieves a $21.4\%$ higher success rate on model \textbf{(2A)} the difference is reduced to $0.2\%$ on model \textbf{(2B)}  . The loss surface visualized in \textbf{(2A)} shows signs of obfuscated gradients. The gradients near the data point are close to zero and therefore provide limited information. Additionally, the loss surface has spiky artifacts which limit the gradient information of a single data point. In contrast, \textbf{(2B)} shows no signs of gradient obfuscation. Consistent with this observation, model \textbf{(2A)} exhibits relatively high $\sigma$ values which results in more smoothing of the loss landscape. Accordingly, for model \textbf{(2B)} the standard deviation $\sigma$ is converging to $0$, as all sampled gradients point towards the same direction and do not provide additional information. 




\section{Conclusion}

In this paper we propose \acf{DSNGD}, an easy-to-implement modification of gradient descent to improve its convergence for non-convex and noisy loss surfaces that scales to high-dimensional optimization tasks. Through our experiments on three different datasets, we demonstrate that this method can be effectively combined with state-of-the-art adversarial attacks to achieve higher success rates. Furthermore, we show that DSNGD-based attacks are substantially more query-efficient than current state-of-the-art attacks. Although the method proved to be effective in our experiments, larger datasets like ImageNet \cite{Deng2009} can be explored. Additionally, the performance in other application areas will be evaluated.


\printbibliography

@book{du2019nonlocal,
  title={Nonlocal Modeling, Analysis, and Computation: Nonlocal Modeling, Analysis, and Computation},
  author={Du, Qiang},
  year={2019},
  publisher={SIAM}
}

@inproceedings{Goodfellow2015,
author = {Goodfellow, Ian and Shlens, Jonathon and Szegedy, Christian},
booktitle = {ICLR},
title = {Explaining and harnessing adversarial examples},
year = {2015}
}

@inproceedings{Madry2018,
title={Towards Deep Learning Models Resistant to Adversarial Attacks},
author={Aleksander Madry and Aleksandar Makelov and Ludwig Schmidt and Dimitris Tsipras and Adrian Vladu},
booktitle={ICLR},
year={2018},
}

@inproceedings{Szegedy2014,
    author    = {Christian Szegedy and
               Wojciech Zaremba and
               Ilya Sutskever and
               Joan Bruna and
               Dumitru Erhan and
               Ian J. Goodfellow and
               Rob Fergus},
  title     = {Intriguing properties of neural networks},
  booktitle = {ICLR},
  year      = {2014},
}

@article{LeCun98,
  title={Gradient-based learning applied to document recognition},
  author={LeCun, Yann and Bottou, L{\'e}on and Bengio, Yoshua and Haffner, Patrick and others},
  journal={Proceedings of the IEEE},
  volume={86},
  number={11},
  pages={2278--2324},
  year={1998},
}

@techreport{krizhevsky2009,
  title={Learning multiple layers of features from tiny images},
  author={Krizhevsky, Alex},
  institution={},
  year={2009},
  DOI={10.1.1.222.9220}
}

@inproceedings{Useato2018,
author    = {Jonathan Uesato and
               Brendan O'Donoghue and
               Pushmeet Kohli and
               A{\"{a}}ron van den Oord},
  title     = {Adversarial Risk and the Dangers of Evaluating Against Weak Attacks},
  booktitle = {Proceedings of the 35th International Conference on Machine Learning, {ICML}},
  volume    = {80},
  pages     = {5032--5041},
  publisher = {{PMLR}},
  year      = {2018},
}

@article{Spall1992,
  author={J. C. {Spall}},
  journal={IEEE Transactions on Automatic Control}, 
  title={Multivariate stochastic approximation using a simultaneous perturbation gradient approximation}, 
  year={1992},
  volume={37},
  number={3},
  pages={332-341},}

@article{Ding2019,
  author    = {Gavin Weiguang Ding and
               Luyu Wang and
               Xiaomeng Jin},
  title     = {advertorch v0.1: An Adversarial Robustness Toolbox based on PyTorch},
  journal   = {CoRR},
  volume    = {abs/1902.07623},
  year      = {2019}
}

@inproceedings{Kaiming2016,
  author = {He, Kaiming and Zhang, Xiangyu and Ren, Shaoqing and Sun, Jian},
  booktitle = {CVPR},
  title = {{Deep Residual Learning for Image Recognition}},
  year = {2016},
}

@inproceedings{Kurakin2018,
author = {Kurakin, Alexey and Boneh, Dan and Tram{\`{e}}r, Florian and Goodfellow, Ian and Kurakin, Alexey and Brain, Google and Papernot, Nicolas and Goodfellow, Ian and Boneh, Dan and McDaniel, Patrick},
booktitle = {ICLR},
title = {Ensemble adversarial training: Attacks and defenses},
year = {2018},
}

@inproceedings{Smith2017,
   author    = {Leslie N. Smith},
  title     = {Cyclical Learning Rates for Training Neural Networks},
  booktitle = {{IEEE} Winter Conference on Applications of Computer Vision},
  pages     = {464--472},
  year      = {2017},
}

@inproceedings{Guo2018,
author = {Guo, Chuan and Rana, Mayank and Cisse, Moustapha and van der Maaten, Laurens},
title = {{Countering Adversarial Images using Input Transformations}},
booktitle = {ICLR},
year = {2018},
}

@inproceedings{Samangouei2018,
  title={Defense-{GAN}: Protecting Classifiers Against Adversarial Attacks Using Generative Models},
  author={Pouya Samangouei and Maya Kabkab and Rama Chellappa},
  booktitle={ICLR},
  year={2018},
}

@inproceedings{Athalye18EOT,
  author    = {Anish Athalye and
               Logan Engstrom and
               Andrew Ilyas and
               Kevin Kwok},
  title     = {Synthesizing Robust Adversarial Examples},
  booktitle = {Proceedings of the 35th International Conference on Machine Learning,
               {ICML}},
  volume    = {80},
  pages     = {284--293},
  publisher = {{PMLR}},
  year      = {2018},
}

@inproceedings{Athalye18Obfuscated,
  author    = {Anish Athalye and
               Nicholas Carlini and
               David A. Wagner},
  title     = {Obfuscated Gradients Give a False Sense of Security: Circumventing Defenses to Adversarial Examples},
  booktitle = {Proceedings of the 35th International Conference on Machine Learning, ICML},
  pages = 	 {274--283},
  year      = {2018},
}

@inproceedings{Oord2016,
  author    = {A{\"{a}}ron van den Oord and
               Sander Dieleman and
               Heiga Zen and
               Karen Simonyan and
               Oriol Vinyals and
               Alex Graves and
               Nal Kalchbrenner and
               Andrew W. Senior and
               Koray Kavukcuoglu},
  title     = {WaveNet: {A} Generative Model for Raw Audio},
  booktitle = {The 9th {ISCA} Speech Synthesis Workshop},
  pages     = {125},
  year      = {2016},
}

@misc{Xiao2017,
  author       = {Han Xiao and Kashif Rasul and Roland Vollgraf},
  title        = {Fashion-MNIST: a Novel Image Dataset for Benchmarking Machine Learning Algorithms},
  volume    = {abs/1708.07747},
  year         = {2017},
  journal = {CoRR}
}

@inproceedings{Lin2020,
  title={Nesterov Accelerated Gradient and Scale Invariance for Adversarial Attacks},
author={Jiadong Lin and Chuanbiao Song and Kun He and Liwei Wang and John E. Hopcroft},
booktitle={ICLR},
year={2020},
}

@inproceedings{tashiro2020,
  author    = {Yusuke Tashiro and
               Yang Song and
               Stefano Ermon},
  title     = {Diversity can be Transferred: Output Diversification for White- and
               Black-box Attacks},
  booktitle = {NeurIPS},
  year      = {2020},
}

@inproceedings{Brendel19,
  author    = {Wieland Brendel and
               Jonas Rauber and
               Matthias K{\"{u}}mmerer and
               Ivan Ustyuzhaninov and
               Matthias Bethge},
  title     = {Accurate, reliable and fast robustness evaluation},
  booktitle = {NeurIPS},
  year      = {2019},
}

@inproceedings{Wong2020,
title={Fast is better than free: Revisiting adversarial training},
author={Eric Wong and Leslie Rice and J. Zico Kolter},
booktitle={ICLR},
year={2020},
}

@inproceedings{Kingma14,
  author    = {Diederik P. Kingma and
               Jimmy Ba},
  title     = {Adam: {A} Method for Stochastic Optimization},
  booktitle = {ICLR},
  year      = {2015},
}

@article{Burke05,
  author    = {James V. Burke and
               Adrian S. Lewis and
               Michael L. Overton},
  title     = {A Robust Gradient Sampling Algorithm for Nonsmooth, Nonconvex Optimization},
  journal   = {{SIAM} J. Optim.},
  volume    = {15},
  number    = {3},
  pages     = {751--779},
  year      = {2005},
  doi       = {10.1137/030601296},
}

@article{Carlini2019,
  author    = {Nicholas Carlini and
               Anish Athalye and
               Nicolas Papernot and
               Wieland Brendel and
               Jonas Rauber and
               Dimitris Tsipras and
               Ian J. Goodfellow and
               Aleksander Madry and
               Alexey Kurakin},
  title     = {On Evaluating Adversarial Robustness},
  journal   = {CoRR},
  volume    = {abs/1902.06705},
  year      = {2019},
  }

@inproceedings{Deng2009,
  title={Imagenet: A large-scale hierarchical image database},
  author={Deng, Jia and Dong, Wei and Socher, Richard and Li, Li-Jia and Li, Kai and Fei-Fei, Li},
  booktitle={CVPR},
  year={2009},
}

@inproceedings{Zhang19,
  author    = {Hongyang Zhang and
               Yaodong Yu and
               Jiantao Jiao and
               Eric P. Xing and
               Laurent El Ghaoui and
               Michael I. Jordan},
  title     = {Theoretically Principled Trade-off between Robustness and Accuracy},
  booktitle = {Proceedings of the 36th International Conference on Machine Learning,
               {ICML}},
  volume    = {97},
  pages     = {7472--7482},
  publisher = {{PMLR}},
  year      = {2019},
  }

@inproceedings{Brendel18,
  author    = {Wieland Brendel and
               Jonas Rauber and
               Matthias Bethge},
  title     = {Decision-Based Adversarial Attacks: Reliable Attacks Against Black-Box
               Machine Learning Models},
  booktitle = {ICLR},
  year      = {2018},
  }

@inproceedings{Mustafa19,
  author    = {Aamir Mustafa and
               Salman H. Khan and
               Munawar Hayat and
               Roland Goecke and
               Jianbing Shen and
               Ling Shao},
  title     = {Adversarial Defense by Restricting the Hidden Space of Deep Neural
               Networks},
  booktitle = {ICCV},
  year      = {2019},
}

@inproceedings{Croce2020,
  author    = {Francesco Croce and
               Matthias Hein},
  title     = {Reliable evaluation of adversarial robustness with an ensemble of diverse parameter-free attacks},
  booktitle = {Proceedings of the 37th International Conference on Machine Learning,
               {ICML}},
  volume    = {119},
  pages     = {2206--2216},
  publisher = {{PMLR}},
  year      = {2020},
  }

@article{rauber2017,
  doi = {10.21105/joss.02607},
  url = {https://doi.org/10.21105/joss.02607},
  year = {2020},
  publisher = {The Open Journal},
  volume = {5},
  number = {53},
  pages = {2607},
  author = {Jonas Rauber and Roland Zimmermann and Matthias Bethge and Wieland Brendel},
  title = {Foolbox Native: Fast adversarial attacks to benchmark the robustness of machine learning models in PyTorch, TensorFlow, and JAX},
  journal = {Journal of Open Source Software}
}

@article{Smilkov17,
  author    = {Daniel Smilkov and
               Nikhil Thorat and
               Been Kim and
               Fernanda B. Vi{\'{e}}gas and
               Martin Wattenberg},
  title     = {SmoothGrad: removing noise by adding noise},
  journal   = {CoRR},
  volume    = {abs/1706.03825},
  year      = {2017},
 }

@article{Wu2018,
  author    = {Lei Wu and
               Zhanxing Zhu and
               Cheng Tai and
               Weinan E},
  title     = {Understanding and Enhancing the Transferability of Adversarial Examples},
  volume    = {abs/1802.09707},
  journal   = {CoRR},
  year      = {2018},
}


\end{document}